%% file: main.tex
\pdfoutput=1
\documentclass[10pt,twocolumn,letterpaper]{article}

\usepackage[pagenumbers]{cvpr}
\usepackage{multirow}
\usepackage{pifont}
\usepackage{makecell}
\usepackage{tabularx}

\input{preamble}
\definecolor{cvprblue}{rgb}{0.21,0.49,0.74}
\usepackage[pagebackref,breaklinks,colorlinks,allcolors=cvprblue]{hyperref}

\def\confName{CVPR}
\def\confYear{2027}

\title{BASE: A Backend-Agnostic Streaming Embodiment Layer for Full-Duplex Conversational Avatars}

\author{Chenye Yang$^{1}$ \quad Jin Jiang$^{1}$ \quad Shengcai Liao$^{1\dagger}$\\
$^{1}$College of Computing and Artificial Intelligence, United Arab Emirates University\\
{\tt\small \{700052938, jinjiang, scliao\}@uaeu.ac.ae}\\
{\small $^{\dagger}$Corresponding author}
}
\usepackage{pgfplots}
\pgfplotsset{compat=1.18}
\usetikzlibrary{patterns}
\usepgfplotslibrary{fillbetween}

\begin{document}

\twocolumn[{%
  \renewcommand\twocolumn[1][]{#1}%
  \maketitle
  \begin{center}
    \includegraphics[width=\textwidth]{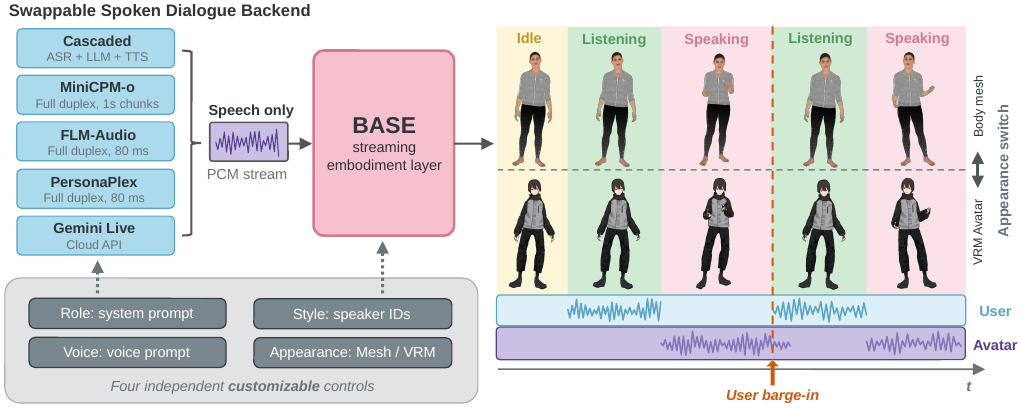}
    \captionof{figure}{\textbf{Overview of our backend-agnostic streaming embodiment layer (BASE).} By decoupling motion generation from internal dialogue states and conditioning exclusively on emitted speech audio, a single embodiment layer turns diverse cascaded, full-duplex and cloud-API spoken dialogue models into customizable, real-time full-body 3D avatars without retraining. Right: the same streamed motion rendered with different appearances. Body mesh texture: Meshcapade sample textures (CC-BY 4.0).}
    \label{fig1}
  \end{center}
}]

\input{sec/0_abstract}
\input{sec/1_intro}
\input{sec/2_relatedwork}
\input{sec/3_method}
\input{sec/4_experiment}
\input{sec/5_conclusion}

{
    \small
    \bibliographystyle{ieeenat_fullname}
    \bibliography{main}
}
\clearpage
\appendix
\setcounter{table}{0}
\setcounter{figure}{0}
\setcounter{equation}{0}
\renewcommand{\thetable}{A\arabic{table}}
\renewcommand{\thefigure}{A\arabic{figure}}
\renewcommand{\theequation}{A\arabic{equation}}
\input{sec/X_suppl.tex}

\end{document}

%% file: sec/0_abstract.tex
\begin{abstract}

Many end-to-end spoken dialogue models can now converse in real time and full duplex, and turning them into 3D avatars requires full-body gestures, listening feedback, and turn-taking. Existing methods couple the motion generator tightly to one dialogue model by reading its internal states or training speech and motion jointly. The motion module must then be retrained whenever the dialogue model changes, cannot be attached to cloud APIs that expose only audio, and entangles role, voice, and motion style. We propose BASE, a backend-agnostic streaming embodiment layer whose only interface to the dialogue model is its emitted speech. First, a conversation state manager driven solely by audio handles speaking, listening, idling, barge-in, and proactive turns on one timeline, while role, voice, motion style, and 3D appearance are set independently. Second, to run an offline 2\,s flow-matching gesture model on live audio, we reformulate inference as sliding-window continuation with overlapped look-ahead ensembling, which removes seam jitter. Third, we show that the remaining streaming gap comes from errors accumulating in the self-generated motion history rather than missing future speech; our Drift-Aware History Perturbation (DAHP) reduces this gap in $\sim$50\,min on one GPU by injecting inference-like drift into ground-truth training histories, and transfers to a second generator. On BEAT2, streaming BASE reaches an FGD of 0.342, on par with the tightly coupled MIBURI (0.333), and viewers prefer its body motion over MIBURI's in 82\% of forced-choice judgments. Without retraining, the same layer drives a cascaded pipeline, three open full-duplex models, and a cloud API in real time.

\end{abstract}

%% file: sec/1_intro.tex
\section{Introduction}
\label{sec:intro}

\begin{table*}[t]
  \caption{\textbf{Comparison of conversational embodied agent frameworks across the coupling spectrum.} Co-speech gesture generators are not connected to a dialogue model; conversational agents read internal dialogue tokens or hidden states, or train a joint multimodal backbone. BASE relies solely on emitted audio, enabling backend swaps across 5 backends and 4-way independent customization at minimal training cost. Quantitative comparisons are given in Table~\ref{tab:quality} and Appendix~\ref{supp:published}. Tool use: whether the agent can call external tools such as web search; BASE inherits it from the backend. $^\dagger$\,Estimated from the paper. $^\ddagger$\,With full-duplex backends; turn-based with the cascade. $^\S$\,Adaptation of a public pretrained generator (sliding-window adaptation $\sim$20\,min, DAHP $\sim$50\,min); the others train from scratch.
}
  \label{tab:framework}
  \centering
  \scriptsize
  \setlength{\tabcolsep}{3pt}
  \renewcommand{\cellalign}{lc}%
  \begin{tabular*}{\textwidth}{@{\extracolsep{\fill}}lllllllll@{}}
    \toprule
    Method & Driving signal & \makecell{Full-\\duplex} & Body scope & Listening & Customization & \makecell{Backend\\swap} & \makecell{Tool\\use} & \makecell{Training\\cost} \\
    \midrule
    \multicolumn{9}{@{}l}{\textit{Co-speech gesture generators (no dialogue model)}} \\
    \quad Offline/Windowed~\cite{liu2024emage,gesturelsm2024,streamtalk2025} & Audio (+ text) & No        & Full body + face & No  & Speaker ID & --          & -- & Per dataset \\
    \quad Streaming~\cite{livegesture2025,gesturefar2025}          & Audio          & Streaming & Full body        & No  & Speaker ID & --          & -- & Per dataset \\
    \midrule
    \multicolumn{9}{@{}l}{\textit{Conversational embodied agents}} \\ \addlinespace[2pt]
    \quad MIBURI~\cite{miburi2025}        & \makecell{Moshi\\internal tokens}     & Yes       & Full body + face     & Yes & Speaker ID        & Retrain    & No & Medium$^\dagger$ \\ \addlinespace[1pt]
    \quad DyaPlex~\cite{dyaplex2026}      & \makecell{PersonaPlex\\hidden states} & Yes       & Body                 & Yes & Voice timbre      & Retrain    & No & 64$\times$H100$^\dagger$ \\ \addlinespace[1pt]
    \quad ELLSA~2~\cite{ellsa2026}        & Joint end-to-end          & Yes       & \makecell{Upper + hands\\+ face} & Yes & Style prompt      & Monolithic & No & \makecell{10.78B,\\multi-stage} \\ \addlinespace[1pt]
    \quad Super Star~\cite{superstar2026} & \makecell{Qwen-Omni \\reply speech}    & Streaming & Body                 & --  & User pref.\ iter. & --         & No & Medium$^\dagger$ \\ \addlinespace[2pt]
    \midrule
    \quad \textbf{BASE (ours)} & \textbf{Emitted audio only} & \textbf{Yes}$^\ddagger$ & \textbf{Full body + face} & \textbf{Yes}
                         & \makecell{\textbf{Role, voice,}\\\textbf{styles, avatar}} & \makecell{\textbf{5 backends,}\\\textbf{no retraining}} & \textbf{Inherited} & \textbf{$\sim$1 GPU-hour}$^\S$ \\
    \bottomrule
  \end{tabular*}
\end{table*}

End-to-end spoken dialogue models (Moshi~\cite{defossez2024moshi}, PersonaPlex~\cite{personaplex2026}, MiniCPM-o~\cite{minicpmo2025}, and FLM-Audio~\cite{flmaudio2025})
can now listen and speak simultaneously, handle mid-sentence interruptions, and reply with natural prosody in real time. Turning these voice models into 3D avatars requires coordinated full-body gestures, responsive listening postures, and natural gaze dynamics~\cite{nyatsanga2023comprehensive,liu2024emage}.

Existing work connects spoken dialogue models to 3D motion by coupling the two tightly, in one of two ways. The first trains motion decoders on frozen internal tokens or hidden states of a specific speech-language model: MIBURI~\cite{miburi2025} reads internal audio-text tokens from Moshi, and DyaPlex~\cite{dyaplex2026} attaches a heavy motion tower to PersonaPlex hidden states. The second group jointly trains speech and non-verbal behavior within a single multimodal model, as in ELLSA~2~\cite{ellsa2026} (speech, face, and gestures) and Moshi-Face~\cite{jiang2026moshiface} (speech and face). Tight coupling brings three bottlenecks. First, dialogue models are replaced every few months, and each new backbone forces a coupled motion module to be retrained from scratch at high compute cost (e.g., $64\times\text{H100}$ GPUs for DyaPlex). Second, proprietary or cloud APIs such as Gemini Live~\cite{geminiliveapi} expose only raw audio, so internal states are unavailable. Third, joint or hidden-state conditioning entangles role, voice, and motion style, so changing one perturbs the others.

In this paper, we advocate complete decoupling to avoid these limitations. We propose \textbf{BASE} (\textbf{B}ackend-\textbf{A}gnostic \textbf{S}treaming \textbf{E}mbodiment), a streaming embodiment layer whose sole interface to the dialogue backend is the emitted speech audio stream (Figure~\ref{fig1}). Our premise is that co-speech gestures are largely governed by vocal prosody, rhythm, and acoustic energy, which the emitted speech carries, and that modern audio-driven motion generators can synthesize realistic movement from it without access to the dialogue model's hidden states. However, realizing a decoupled, real-time full-body avatar entails two major challenges. First, high-fidelity audio-driven gesture models (typically built on diffusion or flow matching~\cite{lipman2023flow,streamtalk2025}) are offline, relying on bidirectional multi-second audio windows and iterative denoising. Second, conversational embodiment goes beyond co-speech gesturing and requires continuous management of listening feedback, idle transitions, interruption damping, long-horizon postural stability, and strict audio-visual synchronization under tight latency budgets.

For the first challenge, sliding a pretrained 2\,s flow-matching generator over live audio fails (FGD 2.01 on BEAT2), and a gap remains after adapting it to sliding windows (0.632, or 0.464 with the seam ensembling below, vs.\ 0.251 with speech in advance). Controlled experiments show that neither 20\,s of future speech nor a longer look-ahead closes this gap, whereas periodically resetting the generator's history to ground truth does. The degradation is therefore exposure bias, i.e., velocity and amplitude errors accumulating in the self-generated history, rather than missing future speech. Guided by this diagnosis, we introduce Drift-Aware History Perturbation (DAHP): during training we perturb the ground-truth history with the error modes seen online (temporal resampling, amplitude scaling, and low-frequency joint noise), decayed to zero at the generation boundary. DAHP needs only the original training data and about 50 minutes on one GPU, lowers streaming FGD (metrics in \S\ref{sec:experiment}) to 0.342, and succeeds where heavier distillation and self-rollout regression fail (Appendix~\ref{supp:alternatives}). Separately, independently sampled windows leave high-frequency jitter at their seams, which we remove by overlapped look-ahead ensembling without training. Both steps transfer to a second generator, GestureLSM~\cite{gesturelsm2024}, lowering its streaming FGD from 1.504 to 1.157.

For the second challenge, a conversation state manager driven only by voice activity on the user's microphone and by the reply audio switches between idling, listening, speaking, barge-in, and proactive turns on one timeline. A backend adapter absorbs output units from sentence-level TTS to 80\,ms full-duplex steps, and each frame is packed with its own audio for synchronization. The system runs on two consumer GPUs (one for the local backend and motion generator, one for rendering) and drives five backends, including a cloud API, without retraining at 1.3--3.0\,s end-to-end latency ($\sim$0.75\,s from the embodiment layer). Our contributions are as follows:

\begin{itemize}
\setlength{\itemsep}{2pt}\setlength{\parskip}{0pt}
    \item \textbf{Backend-Agnostic Embodiment Layer (\S\ref{sec:method_layer}).}
    We design a full-body embodiment layer whose only interface to the dialogue backend is the emitted speech. It handles speaking, listening, barge-in and proactive turns on one timeline, customizes role, voice, motion style and appearance independently, and drives five backends, including a cloud API, without retraining.

    \item \textbf{Seamless Streaming Continuation (\S\ref{sec:method_continuation}).}
    We turn an offline 2\,s flow-matching gesture model into a real-time sliding-window generator and remove window seams with overlapped look-ahead ensembling, which halves torso high-frequency energy (17.1 to 7.9).

    \item \textbf{Drift-Aware History Perturbation (\S\ref{sec:method_perturbation}).}
    We show that the streaming degradation comes from errors accumulating in the self-generated history rather than from missing future speech, and reduce it by training on ground-truth histories perturbed to mimic this drift. In $\sim$50\,min on one GPU, DAHP lowers streaming FGD from 0.464 to 0.342, on par with the tightly coupled MIBURI, and transfers to GestureLSM.
\end{itemize}

%% file: sec/2_relatedwork.tex
\section{Related Work}
\label{sec:relatedwork}

\begin{figure*}[t]
  \centering
  \includegraphics[width=\textwidth]{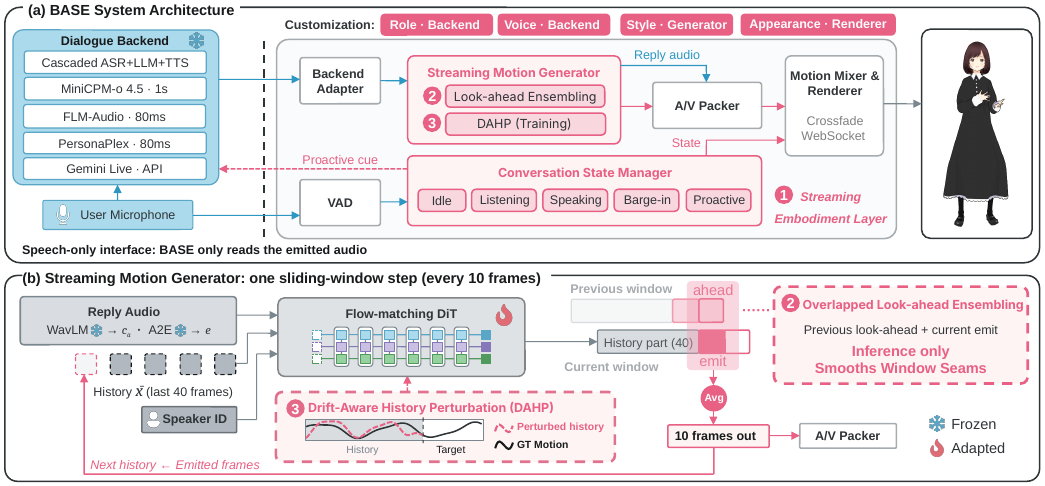}
  \caption{\textbf{Architecture of BASE.} (a)~The embodiment layer~(\ding{172}) reads only the speech audio emitted by any dialogue backend; its conversation state manager driven by VAD and reply-audio events handles idling, listening, barge-in and proactive turns. (b)~Each sliding-window step: a 6-layer part-aware flow-matching DiT performs 8-step continuation (CFG 4) over a 60-frame window, conditioned on WavLM features $c_a$, expressions $e$, the 40-frame self-generated history $\bar{x}_{1:40}$, and a speaker style ID. Overlapped look-ahead ensembling~(\ding{173}, inference only) averages the previous window's look-ahead with the current emitted frames, which are then fed back as history. Drift-Aware History Perturbation~(\ding{174}, DAHP, training only) exposes the generator to inference-like history drift.}
  \label{fig2}
\end{figure*}

\paragraph{Audio-Driven Co-Speech Gesture Generation.}
Co-speech gesture generation has moved from deterministic regression and adversarial models~\cite{ginosar2019learning,yoon2020speech} to VQ-VAE, diffusion, and flow-matching models~\cite{yi2023generating,zhu2023taming,semconflow2025,alexanderson2023listen,liu2024emage,gesturelsm2024,streamtalk2025}, evaluated on motion-capture benchmarks such as BEAT and BEAT2~\cite{liu2022beat,liu2024emage}. Most of these models run offline on full utterances or multi-second bidirectional windows. Rather than designing a new offline architecture, we adapt an offline flow-matching generator for low-latency, long-horizon streaming.

\paragraph{Streaming and Autoregressive Motion Generation.}
Recent streaming gesture models generate motion causally or in chunks, including StreamTalk~\cite{streamtalk2025}, which generates consecutive 2\,s windows, GestureFAR~\cite{gesturefar2025}, LiveGesture~\cite{livegesture2025}, and Accelerated Rolling Diffusion~\cite{ard2025}. A central issue in autoregressive generation is exposure bias: errors in self-generated prefixes accumulate over time. Standard remedies include scheduled sampling~\cite{bengio2015scheduled}, DAgger~\cite{ross2011reduction}, Diffusion Forcing~\cite{chen2024diffusion}, Self Forcing~\cite{selfforcing2025}, and context frame noise injection in interactive world models such as GameNGen~\cite{valevski2024diffusion}. Similarly, smoothing chunk transitions relates to temporal ensembling in robotic action chunking (ACT)~\cite{zhao2023learning}. In contrast, we mitigate streaming exposure bias with DAHP, lightweight boundary-continuous history perturbations tailored to online velocity and amplitude drift, and show that the remedy is not tied to one generator. Moreover, we analyze why heavier distillation and self-rollout schemes fall short in continuous gesture synthesis (Appendix~\ref{supp:alternatives}).

\paragraph{Spoken Dialogue Backends.}
Voice agents have traditionally used ASR--LLM--TTS cascades~\cite{whisper,qwen25,cosyvoice2} that reply sentence by sentence. End-to-end models now listen and speak at once: Moshi~\cite{defossez2024moshi}, PersonaPlex~\cite{personaplex2026} and FLM-Audio~\cite{flmaudio2025} emit audio in 80\,ms frames, and omni-modal models such as MiniCPM-o~\cite{minicpmo2025} in chunks of about 1\,s. Cloud services such as the Gemini Live API~\cite{geminiliveapi} stream speech without exposing any internal state. Their internal representations differ from model to model; the emitted speech is the only interface they share. BASE builds on this interface, and its backend adapter absorbs the different output granularities (\S\ref{sec:exp-backends}).

\paragraph{Dialogue-Driven Embodied Conversational Agents.}
Embodied dialogue agents fall into three groups by how tightly the motion generator is coupled to the dialogue model (Table~\ref{tab:framework}). The first reads frozen internal states: MIBURI~\cite{miburi2025} conditions body and face generation on Moshi's internal audio-text tokens, and DyaPlex~\cite{dyaplex2026} trains a motion tower on PersonaPlex hidden states. Internal features provide look-ahead but tie the avatar to one model checkpoint. The second category trains speech and non-verbal behavior jointly end to end: ELLSA~2~\cite{ellsa2026} optimizes speech, facial expressions, and upper-body gestures within a unified 10.78B-parameter model; Moshi-Face~\cite{jiang2026moshiface} adds facial motion to Moshi. The third covers streaming facial or system avatars~\cite{superstar2026, sentiavatar2025, faceplex2026, chatanyone2025, streamavatar2025}. Closest to our interface is Super Star~\cite{superstar2026}, which trains a causal autoregressive generator that maps streamed reply speech to full-body motion, and refines it with user-preference data. It is built for one speech model and covers only the reply; BASE instead adapts an existing generator, handles listening, barge-in, and proactive turns, runs unchanged on five backends, and analyzes the drift that streaming causes.

%% file: sec/3_method.tex
\section{Method}
\label{sec:method}

Figure~\ref{fig2} illustrates \textbf{BASE}, which has three main components. (a)~The backend-agnostic embodiment layer~(\ding{172}, \S\ref{sec:method_layer}) reads only the reply speech emitted by the dialogue backend and the user's microphone; it manages the conversation state and synchronizes, mixes and renders the motion.
(b)~Inside it, the streaming motion generator encodes the reply audio with two frozen front-ends, a pretrained WavLM~\cite{chen2022wavlm} encoder (acoustic features $c_a$) and the audio-to-expression (A2E) model released with StreamTalk~\cite{streamtalk2025} (facial expressions $e$), and continues the motion in sliding windows whose seams are removed by overlapped look-ahead ensembling~(\ding{173}, \S\ref{sec:method_continuation}). Drift-Aware History Perturbation~(\ding{174}, \S\ref{sec:method_perturbation}) trains it to tolerate its own drift.

\subsection{Backend-Agnostic Embodiment Layer}
\label{sec:method_layer}
BASE never reads the backend's tokens, hidden states or text: everything it needs comes from the emitted speech and the user's microphone (\ding{172} in Figure~\ref{fig2}).
\begin{itemize}
\setlength{\itemsep}{1pt}\setlength{\parskip}{0pt}
    \item \textbf{Speech-Only Interface and Synchronization:} A backend adapter resamples the emitted audio and frames it for the generator, absorbing output units from sentence-level TTS to 80\,ms full-duplex steps and 1\,s chunks. The A/V Packer bundles each body frame with its audio slice and expression $e$, withholding initial playback for a chunk-matched prebuffer (0\,s for 80\,ms steps, 1\,s for 1\,s chunks) plus a 0.2\,s jitter margin so playback never runs dry.
    \item \textbf{Conversation State Manager:} A voice activity detector on the user's microphone and reply-audio events switch between \emph{Idle}, \emph{Listening}, \emph{Speaking}, \emph{Barge-in} and \emph{Proactive}. Speaking motion, seeded with the current idle pose, fades in over 20 frames at speech onset; at speech end or user barge-in it eases over 15 frames into a motion-capture idle clip chosen to match its foot stance. While the user talks, a procedural listening overlay, or the external ELLSA~2~\cite{ellsa2026} listening generator, runs on the same timeline with 10-frame cross-fades. After prolonged mutual silence, synthetic vocal cues are mixed into full-duplex backend inputs to prompt a proactive turn.
    \item \textbf{Four-Way Orthogonal Customization:} Independent specification of \emph{Character Role} (LLM system prompt), \emph{Voice Timbre} (backend audio prompt or preset), \emph{Motion Style} (BEAT2 speaker IDs), and \emph{Visual Appearance} (real-time SMPL-X retargeting to arbitrary VRM avatars, or direct textured body-mesh rendering).
\end{itemize}

\subsection{Seamless Streaming Continuation}
\label{sec:method_continuation}
Our base model is the released StreamTalk~\cite{streamtalk2025} generator, a flow-matching Diffusion Transformer (DiT) conditioned on WavLM~\cite{chen2022wavlm} acoustic features and a discrete speaker style ID. It generates 2\,s windows ($W=60$ frames at 30\,fps) one after another but needs each window's speech in full, so on live audio its motion would lag the speech by a whole window. We adapt it as follows:

\begin{itemize}
\setlength{\itemsep}{1pt}\setlength{\parskip}{0pt}
    \item \textbf{Sliding-Window Continuation:} Within each $W=60$-frame window, the first 40 frames are fixed as previously emitted historical context, and the model generates the remaining 20 frames, emitting the first $N=10$ frames ($0.33\text{ s}$ stride) and retaining the last 10 frames as look-ahead predictions.
    \item \textbf{Sliding-Window Adaptation:} Half of the samples in each training batch become prefix-conditioned completions that generate the last $S \in [6, 20]$ frames; the other half keep the original random masks. (10 epochs, about 20 minutes on one GPU)
    \item \textbf{Overlapped Look-Ahead Ensembling:} To eliminate seam discontinuities between window $k-1$ and window $k$, we blend the newly generated frame $x^{(k)}_f$ with the previous window's look-ahead prediction $x^{(k-1)}_f$ using a linearly decaying weight:
    \begin{equation}
        \hat{x}_f = w_f \, x^{(k-1)}_f + (1 - w_f) \, x^{(k)}_f, \; w_f = 0.5\!\left(1 - \frac{f - f_0}{N}\right)\!,
        \label{eq:ensembling}
    \end{equation}
    where $f_0$ is the first emitted frame of window $k$ and $f \in [f_0, f_0 + N - 1]$. Initial Gaussian noise is anchored to absolute timeline frame indices so consecutive windows share identical stochastic latents on overlapping frames, so that overlapping predictions differ only through the newly arrived audio. The root and torso joints of the first 8 blended frames are then corrected toward the constant-velocity continuation of the last two history frames, with a cosine-faded weight.
    \item \textbf{Real-Time Sampling and Filtering:} We sample with 8 Euler steps and CFG 4 ($97\text{ ms}$ per window within a $333\text{ ms}$ real-time budget), followed by a $1.5\text{ Hz}$ causal low-pass filter applied exclusively to emitted root and spine joints (tightened from 3\,Hz; \S\ref{sec:exp-ablation}).
\end{itemize}

\subsection{Drift-Aware History Perturbation (DAHP)}
\label{sec:method_perturbation}
The remaining streaming gap comes from errors accumulating in the generator's own history rather than from missing future speech (\S\ref{sec:exp-gap}): during live streaming the 40-frame conditioning prefix consists of the model's own predictions, so velocity and amplitude errors compound. DAHP simulates this drift on the conditioning history during training.

\begin{itemize}
\setlength{\itemsep}{1pt}\setlength{\parskip}{0pt}
    \item \textbf{Structured Corruptions Matching Online Drift:} In a second training stage, initialized from the sliding-window model, with probability $p=0.75$ per prefix-conditioned sample, we corrupt the ground-truth prefix $x_{1:P}$ ($P = W - S$, i.e., $40$ at inference) using one of three transformations $T(\cdot)$ that mimic online streaming drift: temporal resampling (by a factor in $[1/1.4, 1.4]$), mean-centered amplitude scaling (by a factor in $[1/1.3, 1.3]$), or Gaussian-smoothed ($4$-frame) joint noise ($\text{std} \le 0.05$).
    \item \textbf{Boundary-Continuous Decay Schedule:} To prevent artificial discontinuities at the conditioning boundary $P$, perturbations decay linearly to zero over the final 8 prefix frames while keeping the target suffix $x_{P+1:W}$ strictly at ground truth:
    \begin{equation}
        \tilde{x}_f = x_f + \left(T(x)_f - x_f\right) \cdot \min\left(1, \frac{P - f}{8}\right), \quad f \le P.
        \label{eq:perturbation}
    \end{equation}
    \item \textbf{Efficiency:} DAHP uses only the original BEAT2 training split ($19.3\text{ hours}$) and completes in $\sim$50\,minutes on a single GPU ($\sim$1.5\% of base training compute). Unlike distribution-matching distillation or self-rollout regression, Eq.~\eqref{eq:perturbation} preserves exact ground-truth target velocity and transition continuity at frame $P$.
\end{itemize}

%% file: sec/4_experiment.tex
\section{Experiments}
\label{sec:experiment}
\newcommand{\lo}{$\downarrow$}
\newcommand{\gt}{$\rightarrow$}

\begin{figure}[t]
  \centering
  \includegraphics[width=\linewidth]{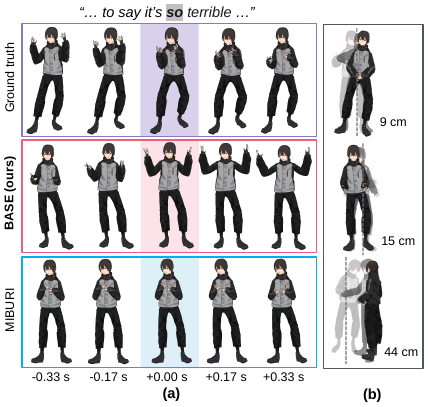}
  \caption{\textbf{Qualitative comparison} on BEAT2 test clips (more examples in Appendix~\ref{supp:qualitative}). (a)~Five frames around stressed words (bold); the shaded column is the onset of ``so'' and offsets are relative to it. Ground truth and BASE gesture with both arms on the emphasis, while MIBURI's hands stay in front of the chest. (b)~Drift in a long clip, rows as in (a): the body at 5\,s and 40\,s (faint) and at 80\,s (solid); the dashed line marks the mean position over the first 10\,s, and the number is the horizontal pelvis displacement at 80\,s (Appendix~\ref{supp:drift}).}
  \label{fig:qualitative}
\end{figure}

\paragraph{Data.}
Motion quality is measured on the 249 BEAT2~\cite{liu2024emage} test clips of the 23 speakers MIBURI~\cite{miburi2025} was trained on. Comparisons with published results use the full official test split (265 clips, all speakers). Our scorer reproduces the speaker-2 FGD reported for GestureLSM~\cite{gesturelsm2024} (0.411 vs.\ 0.409), so our numbers are broadly comparable with the literature. MIBURI's motion for these clips is generated with its released checkpoint and official test script, which feeds Moshi the recorded speech as Mimi audio tokens together with the time-aligned transcript. All motion-quality comparisons (\S\ref{sec:exp-decouple}--\S\ref{sec:exp-ablation}, \S\ref{sec:exp-user}) drive the embodiment layer directly with the recorded test speech, without a dialogue backend; backends are evaluated in \S\ref{sec:exp-backends}.

\paragraph{Metrics.}
FGD measures the distance between the generated and real motion distributions~\cite{liu2024emage}. BeatAlign (BA)~\cite{liu2024emage} and the rhythm alignment $A$ of ELLSA~2~\cite{ellsa2026} measure how well motion follows the speech rhythm ($A=0.5$: no better than unrelated speech). Speed and jerk measure motion energy and smoothness. Torso high-frequency energy (HF) is the share of torso velocity energy above 4\,Hz and exposes window seams. Latency runs from the end of the user's speech to the first avatar frame with sound; FGD and BA of live replies are computed unpaired against BEAT2. In tables, the best value is in bold and the second best underlined.

\paragraph{Implementation.}
Unless stated otherwise, BASE runs with 2\,s windows at 30 fps, a new window every 10 frames with 10 look-ahead frames, 8 flow steps, CFG 4 and a 1.5\,Hz output low-pass. For long live sessions the deployed system additionally applies a weak stance spring that pulls toward the initial stance (Appendix~\ref{supp:drift}), used only in the live experiments \S\ref{sec:exp-backends}--\S\ref{sec:exp-custom}. Numbers are mean $\pm$ std over 3 seeds. Live experiments use five backends without retraining the embodiment layer: PersonaPlex~\cite{personaplex2026}, FLM-Audio~\cite{flmaudio2025}, MiniCPM-o~4.5~\cite{minicpmo2025}, the Gemini Live API~\cite{geminiliveapi}, and a cascade (Whisper~\cite{whisper}, Qwen2.5-7B~\cite{qwen25}, CosyVoice~2~\cite{cosyvoice2}).

\subsection{Does Decoupling Cost Quality?}
\label{sec:exp-decouple}

\begin{table}[t]
  \caption{Motion quality on 249 BEAT2 test clips. Methods in the first group see their speech in advance; the others stream. $^\ddagger$\,Also reads the ground-truth transcript (official all-speaker checkpoint and test protocol). MIBURI reads the dialogue model's internal tokens and is generated from its released checkpoint. Offline: the base model~\cite{streamtalk2025} on consecutive 2\,s windows; Cascade: the same with the first window started after 1\,s of speech, as with our cascade backend. ELLSA~2 standalone: its released audio-to-gesture generator, which produces upper body and hands only. MIBURI and ELLSA~2 are single runs with their released sampling settings. $\downarrow$ lower is better, and $\rightarrow$ closer to the ground-truth value is better.
}
  \label{tab:quality}
  \centering
  \scriptsize
  \setlength{\tabcolsep}{3pt}
  \begin{tabular*}{\columnwidth}{@{\extracolsep{\fill}}lccccc@{}}
    \toprule
    Method & FGD\lo & $A$\gt & BA\gt & Speed\gt & HF\gt \\
    \midrule
    Ground truth                               & --                  & 0.632 & 0.604 & 1.00 & 4.3  \\
    \midrule
    \multicolumn{6}{@{}l}{\textit{Speech in advance}} \\
    \quad Offline, 2\,s windows                & \textbf{0.251\,$\pm$\,0.001} & \textbf{0.661} & \textbf{0.644} & \textbf{1.17} & 36.4 \\
    \quad Cascade, 1\,s early start            & \underline{0.251\,$\pm$\,0.006} & \textbf{0.661} & \underline{0.646}& \underline{1.19} & \underline{36.3} \\
    \quad GestureLSM$^\ddagger$, 4.3\,s windows & 0.723\,$\pm$\,0.002 & \underline{0.524}& 0.696 & 0.65 & \textbf{3.7}  \\
    \midrule
    \multicolumn{6}{@{}l}{\textit{Streaming}} \\
    \quad GestureLSM$^\ddagger$, 0.4\,s sliding & 1.504\,$\pm$\,0.007 & \underline{0.571} & 0.709 & 0.53 & \textbf{3.9}  \\
    \quad ELLSA~2 standalone                   & 3.615               & \underline{0.571} & 0.732 & 1.10 & 50.2 \\
    \quad MIBURI                               & \textbf{0.333}      & 0.520 & \underline{0.670} & \textbf{1.07} & 1.4  \\
    \quad \textbf{BASE (ours)}                 & \underline{0.342\,$\pm$\,0.002} & \textbf{0.635} & \textbf{0.622} & \underline{1.09} & \underline{3.5} \\
    \bottomrule
  \end{tabular*}
\end{table}

Reading only the backend's audio does not reduce motion quality in our comparison (Table~\ref{tab:quality}). With speech available in advance, the same audio-only generator reaches FGD 0.251, below MIBURI (0.333), which is coupled to the dialogue model's internal tokens. When streaming, BASE (FGD 0.342) is on par with MIBURI. GestureLSM~\cite{gesturelsm2024}, an offline generator that additionally reads the ground-truth transcript, reaches only FGD 0.723 with its official all-speaker checkpoint. Run in 0.4\,s sliding windows on the speech received so far, a setting comparable to BASE's, its FGD rises to 1.504 and its motion slows to about half the ground-truth speed. ELLSA 2's standalone generator is far from the data distribution (FGD 3.6). Moreover, BASE is closest to ground truth on both $A$ and BA. On the 171 stressed words of the user-study segments (\S\ref{sec:exp-user}) where ground truth makes a gesture stroke, BASE also strokes on 60\% and MIBURI on 43\% (Figure~\ref{fig:qualitative}a, Appendix~\ref{supp:qualitative}). The offline and cascade rows show high torso HF because 2\,s windows are stitched with visible jumps, and BASE's overlapping windows remove them (from 36.4 to 3.5). Viewers see these differences too: they prefer BASE to both offline windows and MIBURI (\S\ref{sec:exp-user}). Over long replies BASE's pelvis drifts about one third as much as MIBURI's (13.5 vs.\ 43.7\,cm after 80\,s; Figure~\ref{fig:qualitative}b, Appendix~\ref{supp:drift}). The stance spring used in live sessions brings drift into the ground-truth range (8.5 vs.\ 6.1\,cm) at a cost of 0.041 FGD, so it is off in all benchmark comparisons. On the full test split, streaming BASE reaches FGD 0.323 (comparison with published results in Appendix~\ref{supp:published}).

\subsection{Where Does the Streaming Gap Come From?}
\label{sec:exp-gap}

\begin{figure}[t]
  \centering
  \includegraphics[width=\columnwidth]{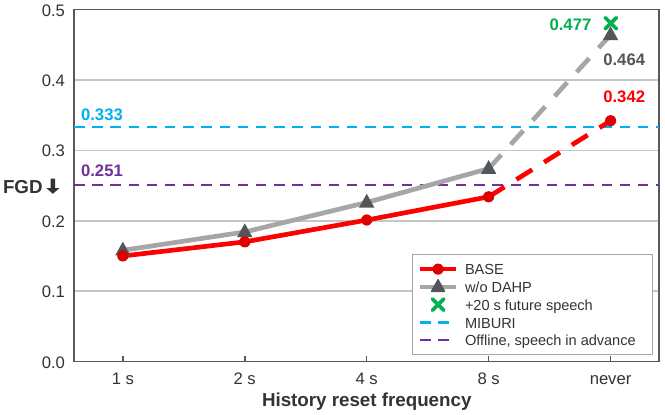}
  \caption{Resetting the model's own history to ground truth every 1\,s cuts FGD from 0.46 to 0.16, while 20\,s of extra future speech does not help. DAHP lowers the whole curve.}
  \label{fig3}
\end{figure}

The FGD cost of streaming may come from missing future speech or from errors accumulating in the model's own history. We separate these two by periodically replacing the history with ground-truth motion while keeping everything else streaming (Figure~\ref{fig3}). Resetting every 1\,s brings FGD from 0.464 to 0.158, even below the offline setting, and the curve rises steadily with the reset interval. Giving the model 20\,s of future speech instead (the WavLM features of each 2\,s window are computed with 20\,s of right context) leaves FGD unchanged (0.477), and a longer look-ahead of 20 frames makes it worse (0.630), since 30 new frames per window exceed the 6--20 the model was adapted for. The gap is therefore exposure bias, not missing context. Because reset frames carry ground-truth information, the 0.158 reached with 1\,s resets is a lower bound rather than an attainable target; the direct evidence is that future speech does not help while DAHP does. DAHP (\S\ref{sec:method_perturbation}) targets exactly this: it lowers the whole curve, most where errors have had longest to accumulate (no reset: from 0.464 to 0.342; every 8\,s: from 0.274 to 0.234). Resets cause jumps at the splice points, so only FGD is meaningful here. This holds for our base model; GestureLSM, trained on 4.3\,s windows, does use future speech (\S\ref{sec:exp-ablation}).

\subsection{Ablation Study}
\label{sec:exp-ablation}

\begin{table}[t]
  \caption{Ablation. Each component helps our base model (0.33\,s stride). Arm HF follows torso HF. All streaming rows use position-anchored noise, the seam correction of \S\ref{sec:method_continuation} and a 3\,Hz torso low-pass unless noted. $\downarrow$ lower is better, and $\rightarrow$ closer to the ground-truth value is better.}
  \label{tab:ablation}
  \centering
  \scriptsize
  \setlength{\tabcolsep}{3pt}
  \begin{tabular*}{\columnwidth}{@{\extracolsep{\fill}}lccccc@{}}
    \toprule
    Configuration & FGD$\downarrow$ & $A$\gt & Speed\gt & Jerk\gt & HF\gt \\
    \midrule
    Ground truth                    & --                  & 0.632 & 1.00 & 58.7  & 4.3  \\
    \midrule
    Pretrained, sliding (5 steps)   & 2.008\,$\pm$\,0.004 & 0.574 & \textbf{0.96} & 120.9 & 15.9 \\
    + sliding-window adaptation     & 0.632\,$\pm$\,0.012 & 0.572 & 0.63 & \underline{62.7}  & 17.1 \\
    + look-ahead ensembling         & 0.518\,$\pm$\,0.004 & \underline{0.617} & \textbf{0.96} & \textbf{59.6}  & 7.9  \\
    + 8 flow steps                  & \underline{0.464\,$\pm$\,0.007} & \underline{0.617} & 1.17 & 72.7  & \underline{6.5}  \\
    + DAHP                          & \textbf{0.342\,$\pm$\,0.002} & \textbf{0.635} & \underline{1.09} & 65.1  & 8.1  \\
    + 1.5\,Hz low-pass (BASE)       & \textbf{0.342\,$\pm$\,0.002} & \textbf{0.635} & \underline{1.09} & 65.5  & \textbf{3.5}  \\
    \bottomrule
  \end{tabular*}
\end{table}

Every component helps our base model (Table~\ref{tab:ablation}). The pretrained model has never continued a given motion prefix, so sliding it directly fails (FGD 2.01); GestureLSM, although trained with a motion seed, also degrades when slid (Table~\ref{tab:quality}). Sliding-window adaptation fixes this but leaves motion slow (0.63$\times$ ground-truth speed). Look-ahead ensembling restores speed and halves seam energy. DAHP is the largest single gain after adaptation (0.464 to 0.342) and the only one that targets the cause found in \S\ref{sec:exp-gap}. Viewers notice this gain: 80\% of judgments favor BASE over BASE without DAHP (\S\ref{sec:exp-user}). Training on self-rollouts via ground-truth regression or distribution-matching distillation fails to improve upon the starting point: regression slows the motion and distillation makes it too fast (Appendix~\ref{supp:alternatives}). Tightening the torso low-pass to 1.5 Hz leaves FGD unchanged while halving residual jitter (HF: 8.1 to 3.5).

The adaptation is not specific to our base model. Applied to GestureLSM~\cite{gesturelsm2024}, which differs in almost every design choice (residual-VQ latents, shortcut flow, 4.3\,s windows, text input), it lowers the streaming FGD from 1.504 to 1.157 ($-23\%$) after 8 minutes of training on one GPU, with DAHP again giving the last step (1.312 to 1.157). The gain is in FGD only: speed stays at the pretrained level and $A$ drops slightly, and about half of GestureLSM's streaming gap is missing future speech, which no change to the history can recover (Appendix~\ref{supp:transfer}).

\subsection{Backend Portability and Latency}
\label{sec:exp-backends}

\begin{table}[t]
  \caption{Five backends driving the same embodiment layer, 10 spoken questions each. Latency: median seconds from the end of the user's speech to the first avatar frame with sound. FGD and BA are unpaired against BEAT2; human speech gives 1.02\,/\,0.60. $\downarrow$ lower is better, and $\rightarrow$ closer to the ground-truth value is better.}
  \label{tab:backends}
  \centering
  \scriptsize
  \setlength{\tabcolsep}{3pt}
  \begin{tabular*}{\columnwidth}{@{\extracolsep{\fill}}lcccccc@{}}
    \toprule
    Backend       & End-to- & Backend & First  & Sync   & \multirow{2}{*}{FGD\lo} & \multirow{2}{*}{BA\gt} \\
    (output unit) & end\lo  & onset\lo   & window\lo & buffer\lo &                      &                     \\
    \midrule
    PersonaPlex (80\,ms)  & \textbf{1.31} & \textbf{0.56} & 0.54 & \underline{0.21}       & 1.22\,$\pm$\,0.05 & \textbf{0.62} \\
    FLM-Audio (80\,ms)    & 2.09 & 1.32 & 0.54 & \underline{0.21}       & \textbf{0.90\,$\pm$\,0.02} & 0.66 \\
    MiniCPM-o 4.5 (1\,s)  & 2.97 & 1.96 & \underline{0.20} & 0.81       & 1.33\,$\pm$\,0.05 & \underline{0.63} \\
    Gemini Live (API)     & 1.61 & 1.24 & \textbf{0.11} & 0.23       & \underline{1.06\,$\pm$\,0.02} & 0.65 \\
    Cascade (TTS)         & \underline{1.47} & \underline{1.23} & 0.24 & \textbf{$\approx$0} & 2.21\,$\pm$\,0.08 & 0.58 \\
    \bottomrule
  \end{tabular*}
\end{table}

The system uses two RTX~5090 GPUs, one for the local backend and the motion generator and one for rendering. Each window (333\,ms of motion) takes 96.7\,ms median and 100.7\,ms at p95 to generate, a real-time factor of 0.29, with 1.9\,GB of GPU memory; the renderer draws a frame in 6.6--12.2\,ms per VRM character. Swapping the backend requires no retraining or changes to the embodiment layer. Only the adapter's prebuffer is set to the backend's output unit (Table~\ref{tab:backends}). The embodiment layer adds about 0.75\,s on the two 80\,ms backends: the first window needs 0.67\,s of
audio for its emitted and look-ahead frames, of which 0.32\,s arrives with the confirmed onset; it is ready 0.54\,s after the onset (0.40\,s for the remaining audio in 80\,ms steps, about 0.1\,s to sample), and the A/V Packer adds a 0.21\,s jitter margin (Appendix~\ref{supp:latency}). Differences between backends lie almost entirely in when the backend starts to speak. Gemini Live delivers each reply faster than real time, so the audio of the first window is already available when the reply starts; its latency is dominated by the cloud onset. For MiniCPM-o, which emits 1\,s chunks, a prebuffer equal to the chunk removes the audio gaps that otherwise hit 23 of 24 replies, at +0.5\,s median latency.

Motion quality from all four full-duplex backends is close to that obtained with human speech (FGD 0.90--1.33 vs.\ 1.02, BA 0.62--0.66 vs.\ 0.60; the reference drives the same layer with BEAT2 test speech cut into 2--8\,s replies). Cascade speech yields worse motion (FGD 2.21), which controlled ablations trace to TTS prosody rather than to BASE (Appendix~\ref{supp:cascade}). Barge-in is independent of the backend: BASE's VAD cancels the reply a median of 0.28\,s after the user starts speaking, on all five backends (robustness to noise and echo in Appendix~\ref{supp:robust}). Listening starts 0.15\,s after the user's speech onset (Appendix~\ref{supp:robust}); proactive turns are shown in the supplementary video. Tool use is inherited from the backend: with Gemini Live's search tool enabled (+0.8\,s onset), the avatar answers time-sensitive questions.

\subsection{Customization}
\label{sec:exp-custom}
The four controls act independently (details in Appendix~\ref{supp:custom}). \emph{Motion style}: one model serves 25 BEAT2 speaker styles; driven with the same speech, the activity statistics of the generated motion rank-correlate with each speaker's own motion capture (Spearman 0.88--0.96), and it lies closer to its own speaker's capture than to the others' (median FGD 1.22 vs.\ 5.18). \emph{Voice} and \emph{role} are backend capabilities that BASE passes through: across 139 PersonaPlex replies (3 roles $\times$ 4 voices), a speaker-verification model identifies the voice at 97\% (chance 25\%), while role adherence is weak in short replies. Style explains 69--86\% of the variance in motion activity, and voice and role under 1\% each. \emph{Appearance}: motion is retargeted in the renderer, so changing the character never touches the embodiment layer (Figure~\ref{fig1}).

\subsection{User Study}
\label{sec:exp-user}
\paragraph{Setup.}
FGD alone does not settle what viewers prefer: offline 2\,s windows have the lowest FGD, while BASE and MIBURI tie.
We therefore ran a two-alternative forced-choice study of body motion. Each question plays one speech clip over two side-by-side renderings and asks which motion is more natural, which fits
the speech better, and which the viewer prefers overall. BASE (benchmark configuration, no stance spring) is compared
with MIBURI, offline 2\,s windows, BASE without DAHP, and ground truth on six 10\,s segments per comparison (24 segments, 23 speakers), each starting 20\,s into a test clip so that streaming errors have accumulated. Character and camera are identical, and both faces are neutral with lips driven by speech loudness, so only body motion differs; this also discards MIBURI's generated face. Sides are balanced and question order is randomized. Twenty university volunteers (aged 20--30) gave informed consent and participated voluntarily for research use; 3 failed an attention check (ground truth vs.\ ground truth from another clip) and were excluded, while all passed the other (ground truth vs.\ pretrained).

\begin{figure}[t]
  \centering
  \includegraphics[width=\columnwidth]{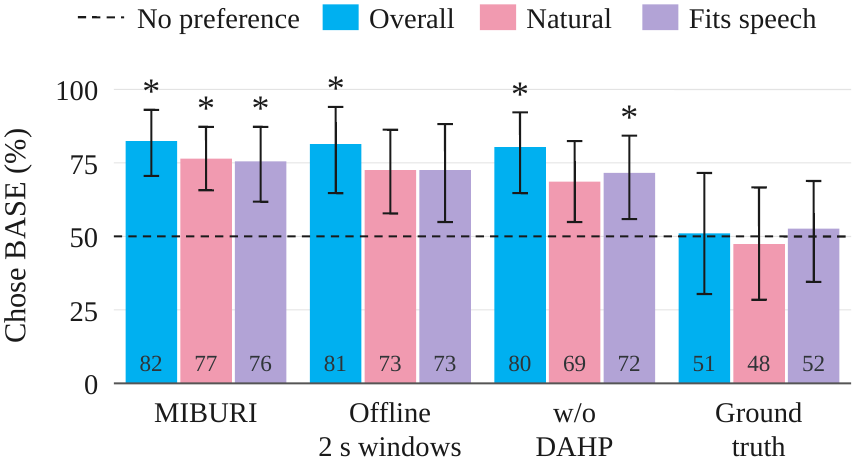}
  \caption{Share of judgments choosing BASE over each alternative (17 participants, 6 clips per comparison). Numbers inside the bars denote the percentage choosing BASE. Error bars: 95\% bootstrap intervals over participants and clips.
  $^*$\,Holm-corrected $p<0.05$.}
  \label{fig:userstudy}
\end{figure}

\paragraph{Results.}
Viewers prefer BASE overall in about 80\% of judgments against MIBURI, offline 2\,s windows and BASE without DAHP (Figure~\ref{fig:userstudy}; $p\le0.008$), and 16 of 17 favor it over MIBURI. The other two questions show the same trend, though three of their six comparisons miss significance after correction. The gain from
DAHP is thus visible to viewers, and FGD misses part of what they judge: offline windows have lower FGD yet lose due to visible seams (torso HF 36.4 vs.\ 3.5), and MIBURI ties on FGD yet loses due to weaker rhythm alignment ($A = 0.520$ vs.\ 0.635). Viewers do not reliably prefer ground truth to BASE. The study is not powered to show equivalence. These are short body-only clips on human speech, so the study does not
measure conversational experience, facial animation or latency.

%% file: sec/5_conclusion.tex
\section{Conclusion and Limitations}
\label{sec:conclusion}

We presented \textbf{BASE}, a streaming embodiment layer whose only interface to a spoken dialogue model is the speech it emits. In our comparison this decoupling does not reduce motion quality: streaming BASE is on par with a generator coupled to the dialogue model's internal tokens, and viewers prefer its body motion in a forced-choice study. We traced the remaining streaming gap to errors accumulating in the generator's own history and reduced it with DAHP, a lightweight drift-aware training scheme that also transfers to another generator. The same layer drives cascaded, full-duplex and cloud-API backends without retraining, with role, voice, motion style, and appearance controlled independently. As dialogue models are replaced every few months, an embodiment layer that only listens to their output can be kept instead of retrained.

\noindent\textbf{Limitations and Future Work.}
Conditioning on audio alone ties motion quality to the prosody of the input speech (Appendix~\ref{supp:cascade}) and precludes purely semantic gestures; transcripts of the emitted audio could supply such cues without coupling to the backend. Turn-taking relies on voice activity detection, so open environments would need echo cancellation or speaker-aware turn detection, and listening behavior is not yet conditioned on what the user says. Finally, exposure bias is reduced but not removed (\S\ref{sec:exp-gap}): long sessions still use a stance spring (Appendix~\ref{supp:drift}), and most of the remaining latency waits for first-window audio.

%% file: sec/X_suppl.tex
\section{Listening and Barge-in Robustness}
\label{supp:robust}

\begin{table}[t]
  \caption{Live BASE with PersonaPlex, 5 trials per condition, a fresh backend session per condition. Disturbances are applied while the avatar replies; echo gain is relative to the question level.}
  \label{tab:robust}
  \centering
  \scriptsize
  \setlength{\tabcolsep}{4pt}
  \begin{tabularx}{\columnwidth}{@{}lc>{\raggedright\arraybackslash}X@{}}
    \toprule
    Condition & Replies cut & Notes \\
    \midrule
    Quiet (control)                          & 0/5           & listening onset 0.15\,s, covers 93\% of the question \\
    White noise $-40$/$-30$/$-20$\,dBFS      & 0/15          & onset 0.15\,s \\
    Own-voice echo $-30$/$-20$/$-14$\,dB     & 2/5, 5/5, 3/5 & false barge-ins \\
    Babble $-35$/$-25$\,dBFS                 & 2/5, 3/5      & false barge-ins \\
    Backchannel (``yeah''), 2\,s into reply  & 5/5           & stop 0.59\,s after onset \\
    Interruption 0/$-12$\,dB                 & 10/10         & stop 0.28\,s after onset (median) \\
    Pause 0.5/1.0/1.5\,s in a question       & --            & listening held 5/5, 0/5, 0/5; backend barged in on 0/5, 0/5, 1/5 \\
    \bottomrule
  \end{tabularx}
\end{table}

The VAD does not separate the user from other speech, so echo, babble and backchannels stop the
avatar; pauses longer than the 700\,ms end-of-turn threshold end the listening state. Practical deployments thus require headphones or acoustic echo cancellation, while speaker-aware turn detection remains future work.

\section{Why Cascade Speech Drives Worse Motion}
\label{supp:cascade}

\begin{table}[t]
  \caption{The cascade's reply texts re-synthesized under controlled variants (3 seeds), same motion layer. $\downarrow$ lower is better, and $\rightarrow$ closer to the reference is better.}
  \label{tab:cascade}
  \centering
  \scriptsize
  \setlength{\tabcolsep}{3pt}
  \begin{tabular*}{\columnwidth}{@{\extracolsep{\fill}}lccccc@{}}
    \toprule
    \multirow{2}{*}{Speech} & FGD & BA & Pauses & Onsets & Pitch std \\
                            & \lo & \gt  & /min\gt   & /s\gt   & (st)\gt      \\
    \midrule
    Human speech (reference)         & 1.02              & 0.60 & 9.5  & 4.7 & 4.0 \\
    TTS female, trimmed (deployed)   & 1.98\,$\pm$\,0.09 & \underline{0.59} & \underline{20.4} & 3.8 & \textbf{3.9} \\
    TTS male, trimmed                & 2.04\,$\pm$\,0.07 & \underline{0.59} & 21.5 & \underline{4.1} & \textbf{3.9} \\
    TTS clone of test speaker, trimmed & 2.11\,$\pm$\,0.03 & \textbf{0.60} & \textbf{16.6} & \textbf{4.3} & \underline{3.6} \\
    TTS female, untrimmed            & \textbf{1.69\,$\pm$\,0.10} & 0.63 & 33.8 & 3.1 & 5.6 \\
    TTS female, whole reply at once  & \underline{1.86\,$\pm$\,0.01} & 0.57 & 26.1 & 3.4 & 5.1 \\
    \bottomrule
  \end{tabular*}
\end{table}

Neither the voice (the test speaker's own cloned voice is no better) nor sentence-by-sentence synthesis
explains the gap; trimming silence accounts for about 0.3 FGD, while the remainder stems from TTS prosody (twice as many pauses and fewer onsets).

\section{Customization Details}
\label{supp:custom}

\paragraph{Motion style.} For each of the 25 BEAT2 speaker styles we generate the same speech (6 clips from 6
different speakers, 3 seeds) and compare the motion with each speaker's own motion capture. Across styles, the
activity statistics of the generated motion rank-correlate with those of the capture: Spearman 0.88 for wrist
speed, 0.89 for upper-body speed and 0.96 for stance drift. From these statistics a speaker is identified at
43\% top-1 (chance 4\%). In the latent space of the FGD encoder, the generated motion lies at FGD 1.22 from its
own speaker's capture and 5.18 from the others' (medians).

\paragraph{Voice.} Across 139 PersonaPlex replies (3 roles $\times$ 4 voices), a speaker-verification model
identifies the voice at 97\% (chance 25\%), and at 96\% when trained only on replies of the other roles, so the
voice does not depend on the role.

\paragraph{Independence.} Style explains 69--86\% of the variance in motion activity, and voice and role
explain under 1\% each.

\paragraph{Role adherence.} An LLM judge picks the persona of each PersonaPlex reply from three descriptions at
46\% (chance 33\%; teacher 84\%, chef 27\%, meditation coach 27\%), and at 44--49\% within each voice, so we
find no effect of voice on persona. Chef and coach replies are mostly judged as teacher (27\% each), so the role
prompt has little effect on short replies. A cascade backend with the same prompts reaches 50\%: short spoken
replies carry little persona. BASE only passes the prompt to the backend.

\section{Latency Details and Compute}
\label{supp:latency}

\begin{itemize}
  \item Embodiment-layer latency: breakdown in \S\ref{sec:exp-backends}. In Table~\ref{tab:backends}, the first-window column is the audio wait plus sampling, and the sync buffer is the jitter margin of the A/V Packer (\S\ref{sec:method_layer}).
  \item MiniCPM-o without a prebuffer: median 2.48\,s, but 23/24 replies break for 140\,ms about 0.8\,s in;
        with a 1\,s prebuffer 2.97\,s and no breaks (0/23).
  \item Cascade: about half the replies break for 120\,ms at the first sentence boundary; a short prebuffer
        would remove this.
  \item GPU memory: motion layer 1.9\,GB (2.7\,GB serving process), renderer 3.5\,GB; local backends 17--24\,GB; Gemini Live runs in the cloud and uses no local GPU memory.
        Seven VRM characters render at 81--151\,fps with 1.0--1.2\,GB each.
  \item Gemini Live: replies arrive faster than real time, so the first window's audio is already available when the reply starts (first window 0.11\,s, synchronization buffer 0.23\,s); latency is dominated by the cloud onset (1.24\,s, including the network round trip). Settings: model gemini-3.8-live, prebuilt voice Puck, search tool off, the teacher prompt of Appendix~\ref{supp:custom}, the API's own voice activity detection; the FGD and BA of Table~\ref{tab:backends} use all 20 replies (223\,s) recorded for the 10 questions.
\end{itemize}

\section{Pelvis Drift and the Stance Spring}
\label{supp:drift}

\paragraph{Drift.} A streaming generator has no notion of where it started, so over long replies the pelvis
wanders and the feet slowly close in. We measure drift as the horizontal pelvis displacement from each clip's first
10\,s, at 20, 40, 60 and 80\,s, over the clips long enough for each horizon (Table~\ref{tab:drift}). BASE drifts
about one third as much as MIBURI: 13.5 vs.\ 43.7\,cm after 80\,s, against 6.1\,cm for ground truth. Only 20 clips
reach 80\,s, so that column is noisier; ground truth itself drifts less at 80\,s than at 60\,s.

\paragraph{Stance spring.} For long live sessions the deployed system adds a weak, training-free stance spring.
After each window (stride $n = 10$ frames) the horizontal root position $\bar r$ is pulled toward its starting
position $r_\text{home}$:
\begin{equation}
  \Delta r = \bigl(1 - e^{-n/(\tau\cdot\text{fps})}\bigr)\cdot \operatorname{shrink}_\delta(\bar r - r_\text{home}),
  \label{eq:spring}
\end{equation}
where $\operatorname{shrink}_\delta$ leaves offsets below the dead zone $\delta_\text{root} = 8$\,cm untouched
(ground-truth motion stays within about 10\,cm) and $\tau = 3$\,s, i.e.\ about 10\% of the remaining offset per
window. The legs are treated the same way: once the left--right foot vector deviates by more than
$\delta_\text{foot} = 2$\,cm from the stance the reply had 2\,s in, the leg joints are pulled back with the same time
constant. The corrected frames are written back into the history buffer, so later windows continue from the
corrected stance instead of fighting it.

\paragraph{Effect and cost.} The spring brings drift into the ground-truth range (8.5--9.1\,cm at 60--80\,s), while
BA, speed, high-frequency energy and stance width stay unchanged (Table~\ref{tab:drift}). It costs 0.041 FGD: FGD
compares clips of at most 90\,s against the data distribution and penalizes the pull back to a home stance, whereas
users in conversations of several minutes see the drift. We therefore leave it off in all benchmark comparisons
(\S\ref{sec:exp-decouple}--\S\ref{sec:exp-ablation}, \S\ref{sec:exp-user}) and on in the live experiments (\S\ref{sec:exp-backends}--\S\ref{sec:exp-custom}) and in Appendix~\ref{supp:cascade}; the FGD and BA of
Table~\ref{tab:backends} are measured with it.

\begin{table}[h]
  \centering
  \caption{Pelvis drift and the stance spring (249 BEAT2 test clips, 3 seeds for BASE). Drift is the horizontal
  pelvis displacement from each clip's first 10\,s, over clips long enough for each horizon (249 clips at 20\,s,
  117 at 60\,s, 20 at 80\,s). MIBURI's BA, speed and HF are from Table~\ref{tab:quality}. $\downarrow$ lower is better,
  and $\rightarrow$ closer to the ground-truth value is better; bold marks the best value apart from ground truth.}
  \label{tab:drift}
  \setlength{\tabcolsep}{3pt}
  \resizebox{\linewidth}{!}{%
  \begin{tabular}{lcccc cccc}
    \toprule
    & & & & & \multicolumn{4}{c}{Drift (cm) after $\rightarrow$} \\
    \cmidrule(lr){6-9}
    Method & FGD$\downarrow$ & BA$\rightarrow$ & Speed$\rightarrow$ & HF$\rightarrow$
           & 20\,s & 40\,s & 60\,s & 80\,s \\
    \midrule
    Ground truth      & --                  & 0.604 & 1.00 & 4.3 & 6.0 & 7.3 & 8.2 & 6.1 \\
    MIBURI            & \textbf{0.333}      & 0.670 & \textbf{1.07} & 1.4 & 24.3 & 34.1 & 32.9 & 43.7 \\
    BASE              & 0.342\,$\pm$\,0.002 & 0.622 & 1.09 & 3.5 & 9.1 & 12.5 & 13.5 & 13.5 \\
    \quad + Stance spring & 0.383\,$\pm$\,0.004 & \textbf{0.620} & 1.08 & \textbf{3.8}
                      & \textbf{7.4} & \textbf{8.8} & \textbf{9.1} & \textbf{8.5} \\
    \bottomrule
  \end{tabular}}
\end{table}

\section{Comparison with Published Results}
\label{supp:published}

\begin{table}[t]
  \caption{FGD on the full BEAT2 test split (265 clips, all speakers). Reported numbers are copied from the papers; re-run rows use our scorer, which gives 0.240 for the released StreamTalk model that its paper reports at 0.293, so differences of a few hundredths across papers are within protocol variation.}
  \label{tab:published}
  \centering
  \scriptsize
  \setlength{\tabcolsep}{4pt}
  \begin{tabular*}{\columnwidth}{@{\extracolsep{\fill}}lllc@{}}
    \toprule
    Method & Source & Speech & FGD$\downarrow$ \\
    \midrule
    CaMN~\cite{liu2022beat}                 & reported & offline         & 0.512 \\
    EMAGE~\cite{liu2024emage}               & reported & offline         & 0.692 \\
    RAG-Gesture~\cite{mughal2025retrieving} & reported & offline         & 0.487 \\
    SemTalk~\cite{zhang2025semtalk}         & reported & offline         & 0.356 \\
    StreamTalk~\cite{streamtalk2025}        & reported & 2\,s windows    & 0.293 \\
    StreamTalk, released                    & re-run   & 2\,s windows    & 0.240\,$\pm$\,0.002 \\
    GestureLSM~\cite{gesturelsm2024}        & re-run   & 4.3\,s windows  & 0.708\,$\pm$\,0.002 \\
    \midrule
    GestureLSM                              & re-run   & 0.4\,s sliding  & 1.513\,$\pm$\,0.005 \\
    \textbf{BASE (ours)}                    & re-run   & 0.33\,s sliding & 0.323\,$\pm$\,0.001 \\
    \bottomrule
  \end{tabular*}
\end{table}

Under live streaming, BASE (0.323 FGD) matches or outperforms the published offline methods, although all of them see
their speech in advance; only the 2\,s-window model it builds on is clearly better. LiveGesture~\cite{livegesture2025} reports results for speaker 2 only and is therefore not listed.

\section{Transferring the Streaming Adaptation to GestureLSM}
\label{supp:transfer}

To test whether our streaming adaptation is specific to the base model, we apply it to
GestureLSM~\cite{gesturelsm2024}, whose design differs throughout: residual-VQ motion latents (4 frames per latent),
a latent shortcut flow, 4.3\,s windows seeded with the last 16 generated frames, and text as a second input.
We start from its official all-speaker checkpoint and keep its test protocol (ground-truth transcript and root
translation, first seed from ground truth; latents are decoded once per clip, as in its own test).

\paragraph{Streaming setting.} A new window starts every 3 latents (12 frames, 0.4\,s). Audio and words are given
up to 10 frames past the emitted ones, as for BASE, and zeroed after; the first 3 new latents are emitted.

\paragraph{Adaptation.} The three components map one to one:
\begin{itemize}
  \item \emph{Sliding-window adaptation.} With probability 0.5 a training window's audio and words are zeroed after
        $16 + 4k + 10$ frames, $k \in \{1,\dots,6\}$, which reproduces what a sliding window receives. The
        flow-matching and shortcut-consistency losses are computed only on latents before the cut (see below).
        Three epochs on the all-speaker training split (97k windows), Adam, learning rate $2\times10^{-5}$,
        batch 128: about 8 minutes on one RTX 5090.
  \item \emph{Look-ahead ensembling.} Eq.~\eqref{eq:ensembling} on the emitted latents ($N = 3$), with initial noise anchored to
        absolute latent positions.
  \item \emph{DAHP.} The same three corruptions with the same ranges and $p = 0.75$
        (\S\ref{sec:method_perturbation}), applied to the 16 seed frames in pose space before they are encoded into seed latents and
        eased back to the clean motion over the last 8 seed frames; the target latents stay clean.
\end{itemize}

\begin{table}[t]
  \caption{Streaming adaptation transferred to GestureLSM (249 BEAT2 test clips, 3 seeds; 2 for the diagnostic row). Reference rows see more speech than streaming allows. $\downarrow$ lower is better, and $\rightarrow$ closer to the ground-truth value is better.}
  \label{tab:transfer}
  \centering
  \scriptsize
  \setlength{\tabcolsep}{2pt}
  \begin{tabular*}{\columnwidth}{@{\extracolsep{\fill}}lcccccc@{}}
    \toprule
    Configuration & FGD$\downarrow$ & $A$\gt & BA\gt & Speed\gt & Jerk\gt & HF\gt \\
    \midrule
    Ground truth                  & --                  & 0.632 & 0.604 & 1.00 & 58.7 & 4.3 \\
    \midrule
    \multicolumn{7}{@{}l}{\textit{Reference}} \\
    Offline, 4.3\,s windows       & 0.723\,$\pm$\,0.002 & 0.524 & 0.696 & 0.65 & 73.9 & 3.7 \\
    Sliding, full future speech   & 1.075\,$\pm$\,0.004 & 0.548 & 0.723 & 0.52 & 66.4 & 3.7 \\
    \midrule
    \multicolumn{7}{@{}l}{\textit{Streaming, 0.4\,s sliding}} \\
    Pretrained model              & 1.504\,$\pm$\,0.007 & \textbf{0.571} & \underline{0.709} & 0.53 & \textbf{65.5} & \textbf{3.9} \\
    + sliding-window adapt.       & 1.951\,$\pm$\,0.026 & \underline{0.540} & 0.716 & \underline{0.56} & 67.2 & \underline{3.7} \\
    + look-ahead ensembling       & \underline{1.312}\,$\pm$\,0.007 & 0.530 & \textbf{0.689} & \textbf{0.61} & 70.8 & \underline{3.7} \\
    + DAHP                        & \textbf{1.157}\,$\pm$\,0.005 & 0.532 & \textbf{0.689} & 0.55 & \underline{66.1} & \underline{3.7} \\
    \bottomrule
  \end{tabular*}
\end{table}

\paragraph{Results.} The full adaptation lowers streaming FGD from 1.504 to 1.157 ($-23\%$;
Table~\ref{tab:transfer}). Sliding-window adaptation alone does not help (1.951); the gain appears with
look-ahead ensembling, and DAHP again gives the last step (1.312 $\rightarrow$ 1.157). Unlike for our base
model, HF stays at 3.7 throughout: the convolutional RVQ decoder already smooths latent seams, so ensembling
acts on FGD only. Speed stays at the pretrained level (0.55 vs.\ 0.53), $A$ drops slightly (0.571 to 0.532)
but stays above the offline model's 0.524, and BA moves toward ground truth.

\paragraph{Where the remaining gap comes from.} Giving the sliding windows their full future speech lowers FGD
from 1.504 to 1.075, whereas 20\,s of future speech leaves our base model unchanged (\S\ref{sec:exp-gap}). GestureLSM,
trained on 4.3\,s windows, therefore relies on future speech, and about half of its streaming gap
($0.43$ of $0.78$) is missing context that no change to the history can recover. Our adaptation recovers
0.35, about as much as the rest of the gap (1.075 vs.\ 0.723).

\paragraph{The loss must be masked.} In a first version the loss was also computed on latents whose speech had
been zeroed. The model then learned to stop moving without speech: streaming motion slowed to 0.43 of
ground-truth speed (FGD 1.907 after sliding-window adaptation, 1.325 with ensembling), and DAHP had no effect
(1.324). Restricting the loss to latents whose speech has arrived keeps speed at 0.55--0.61 and lets DAHP
work. Our base model does not need this mask because its sliding-window adaptation
(\S\ref{sec:method_continuation}) never zeroes audio.

\begin{figure*}[t]
  \centering
  \includegraphics[width=\textwidth]{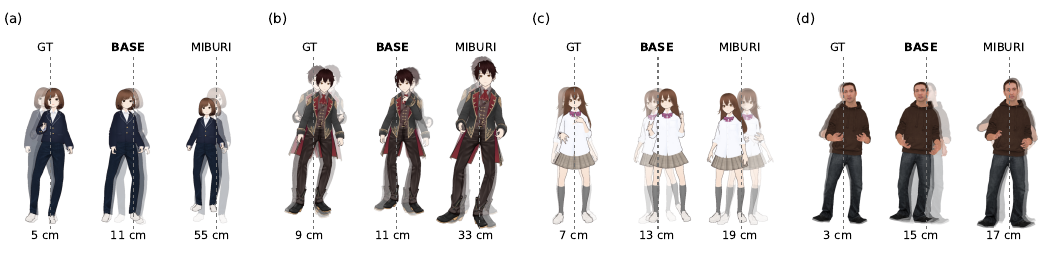}
  \caption{\textbf{Drift on four further clips and characters.} As in Figure~\ref{fig:qualitative}b: the body at 5\,s and
40\,s (faint) and at 80\,s (solid); the dashed line marks the mean position over the first 10\,s, and the
number is the horizontal pelvis displacement at 80\,s.}
\label{fig:supp_drift}
\end{figure*}

\section{Heavier Alternatives to DAHP}
\label{supp:alternatives}

Before DAHP we tried the two standard remedies for exposure bias, both of which train on the
model's own rollouts: regression toward ground truth from self-generated history, and distribution-matching
distillation (DMD) on self-rollouts in the style of Self Forcing~\cite{selfforcing2025}. Neither improved on its starting
point. These runs predate the final configuration, so they were screened on a fixed subset of 67 test clips with one
seed and CFG~3. Each row is compared with its own reference, evaluated under the same settings; seed noise on this
subset is about $\pm0.02$ FGD.

\paragraph{Self-rollout regression.} A chain of consecutive training windows is rolled out without gradients for
2--5 windows (1--2\,s) on the model's own history, as at inference. The next window is then trained with this
self-generated history as its prefix and the ground-truth continuation as its target. The offset between the
history's last frame and the ground truth is added to the target and faded out over 60 frames, since a shorter fade
taught the model fast jumps. A second variant (round 2) mixed 32 online own-history rows into each batch of DAHP training.

\paragraph{Self-rollout DMD.} The student generates whole windows on its own history, and only the newly generated
frames receive gradients. The frozen offline model acts as the teacher and judges the 2\,s window (history plus new
frames) at a random noise level, and a fake critic is trained on the student's windows. The student moves along the
difference of their predictions, with no ground-truth target; a small supervised flow-matching term keeps it close
to the data. Five rounds addressed the failure of the previous one:
(1)~4-step student without CFG;
(2)~an 8-step student and a second ``seam'' teacher that sees the 40 history frames clean, as at inference;
(3)~persistent streams, in which each batch row walks through a whole training clip on its own history, plus EMA;
(4)~starting from the DAHP model, a clip-level speed term that pulls a stream's running speed
ratio back into $1\pm0.1$;
(5)~that speed term on body joints only, with rollouts at the inference stride of 10 frames, including look-ahead
ensembling.
Each round trained 2000--4000 steps (1--2\,h on one GPU) with student, teacher(s) and critic(s) in memory.

\begin{table}[t]
  \caption{Self-rollout training vs.\ DAHP (67-clip screen, one seed, CFG~3; the last row uses all 249 clips, hence its different Ref.). Start: the model training began from; Ref.: its FGD under the same settings. Best: the best checkpoint on this screen (optimistic); Last: the final checkpoint(s). Speed: range over checkpoints (ground truth 1.00).}
  \label{tab:alternatives}
  \centering
  \scriptsize
  \setlength{\tabcolsep}{3pt}
  \renewcommand{\cellalign}{lc}%
  \begin{tabular*}{\columnwidth}{@{\extracolsep{\fill}}llcccc@{}}
    \toprule
    Method & Start & Ref.\ FGD & Best & Last & Speed \\
    \midrule
    \makecell{Self-rollout\\regression}      & sliding, 5 steps & 0.651 & 0.590 & 0.624         & 0.69--0.83 \\ \addlinespace[2pt]
    DMD (1)                                  & sliding, 8 steps & 0.538 & 0.576 & 0.710         & 1.31--1.57 \\ \addlinespace[2pt]
    \makecell{DMD (2)\\+ seam teacher}       & sliding, 8 steps & 0.538 & 0.532 & 0.590\,/\,0.616 & 1.27--1.54 \\ \addlinespace[2pt]
    \makecell{DMD (3)\\+ persistent streams} & sliding, 8 steps & 0.538 & 0.528 & 0.539\,/\,0.592 & 1.26--1.52 \\ \addlinespace[2pt]
    \textbf{DAHP}                            & sliding, 8 steps & 0.538 & \textbf{0.454} & \textbf{0.463} & 0.94--1.08 \\
    \midrule
    \makecell{DMD (4)\\+ speed term}         & DAHP             & 0.437 & 0.449 & 0.478\,/\,0.507 & 1.13--1.20 \\ \addlinespace[2pt]
    \makecell{DMD (5)\\+ body speed}         & DAHP             & 0.437 & 0.457 & 0.473\,/\,0.538 & 1.03--1.12 \\ \addlinespace[2pt]
    \makecell{Self-rollout\\regression (2)}  & DAHP             & 0.355 & 0.411 & 0.411         & 0.87--0.94 \\
    \bottomrule
  \end{tabular*}
\end{table}

\paragraph{Findings.}
\emph{Regression toward ground truth from self-generated history drains motion energy.} Speed falls from 0.85 to
0.69--0.83 in the first variant and from 1.01 to 0.87--0.94 in the second. Once the history has drifted, the
ground-truth continuation no longer continues it; the regression target then averages incompatible continuations,
and the model learns to move less. A third variant replaced the reconstruction loss on cross-speaker rollouts with a
stance loss alone and collapsed (FGD 8.6--13.9).

\emph{DMD removes seams but makes motion too fast.} From round 2 on, torso HF falls to about 1 and BeatAlign rises
(0.62--0.65 vs.\ 0.61), but speed overshoots to 1.3--1.6 and keeps increasing beyond the rollout horizon seen in
training: in round 2, 1.06 within the first 3\,s of a clip and 1.31--1.34 after. The teacher itself is not too fast
(0.93--1.09 offline at CFG 1--3), so the overshoot comes from the student: a critic that judges 2\,s windows does not
constrain statistics that build up over a clip. An explicit speed term restores speed (round 5: 1.03--1.06), but FGD
stays above its starting point (0.457--0.538 vs.\ 0.437), so speed was not the whole gap.

\emph{DAHP is simpler and better.} It simulates inference-time drift directly on ground-truth histories, so it
needs no rollouts, teacher or critic, and it keeps exact ground-truth targets and transition continuity at the
boundary (Eq.~\eqref{eq:perturbation}). It is the only variant that improved on
its starting point (0.538 $\rightarrow$ 0.454 on this screen, 0.464 $\rightarrow$ 0.342 on the full benchmark at
CFG~4) while keeping speed near ground truth and flat over time. DMD applied on top of it did not help further.

\begin{figure*}[t]
  \centering
  \includegraphics[width=\textwidth]{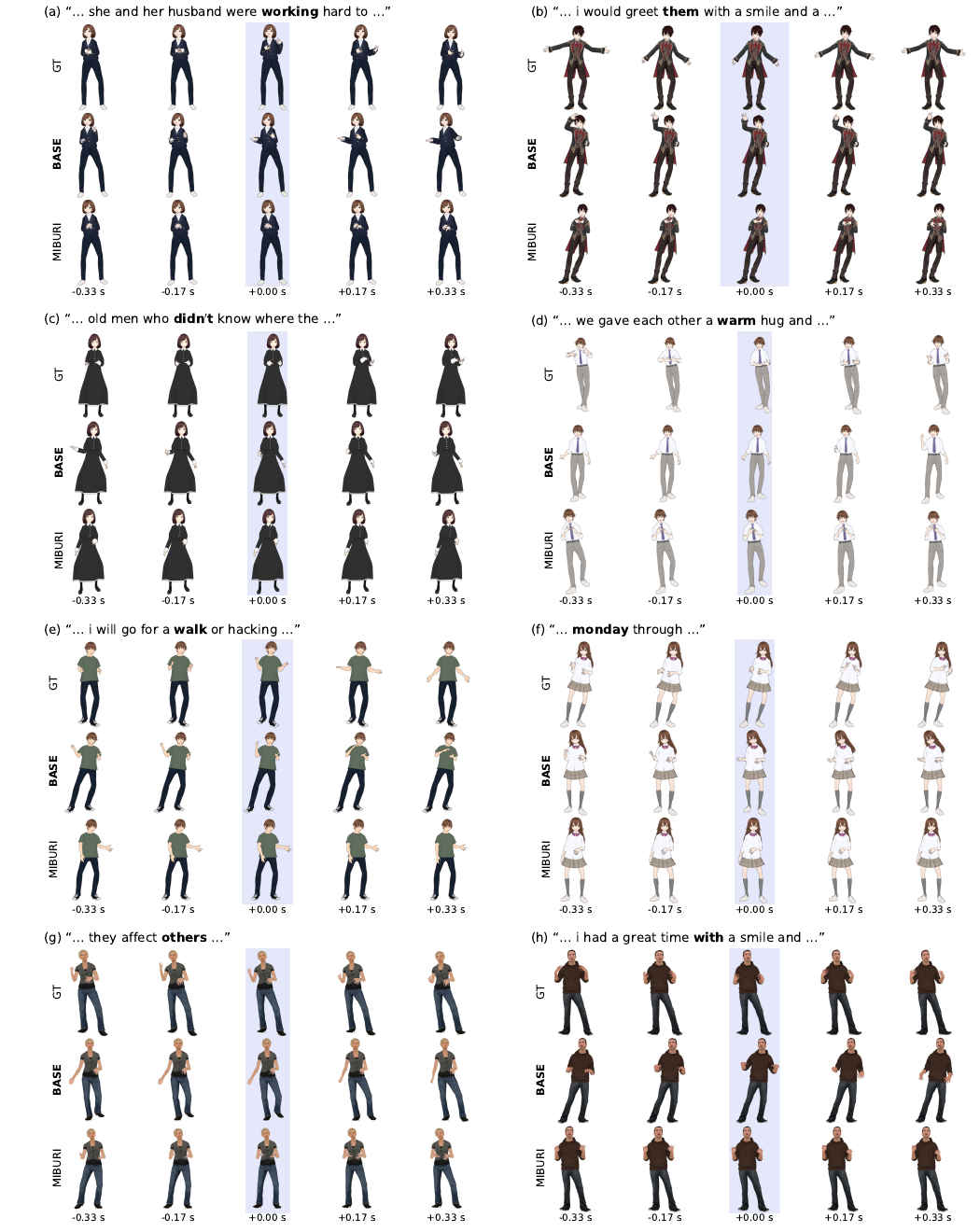}
  \caption{\textbf{Gesture strokes on eight further characters.} Rows and columns as in Figure~\ref{fig:qualitative}a:
five frames around a stressed word (bold; shaded column: its loudness peak). Selection and statistics in the
text. Characters: (c) pixiv VRoid Project; (g, h) Microsoft Rocketbox.}
  \label{fig:supp_strokes}
\end{figure*}

\section{Additional Qualitative Results}
\label{supp:qualitative}

Figures~\ref{fig:supp_drift} and~\ref{fig:supp_strokes} extend Figure~\ref{fig:qualitative} to the eight other VRM characters of our system. Each example uses a different character and speaker. The motion is retargeted in the renderer, so no part of the embodiment layer changes between characters (\S\ref{sec:exp-custom}). All methods are rendered with the same character and camera, and BASE runs without the stance spring. In Figure~\ref{fig:supp_drift}, the four clips are those whose drift at 40\,s and 80\,s is closest to the test-set means of Table~\ref{tab:drift}, one per speaker and excluding the speaker of Figure~\ref{fig:qualitative}b. They are not selected for a large gap: in clip (d) BASE and MIBURI drift about equally (15 vs.\ 17\,cm). The view is frontal, so displacement in depth appears only as a change of size. In Figure~\ref{fig:supp_strokes}, we take the stressed words of the 24 user-study segments, i.e.\ the words whose loudness peak lies in the top quarter of their segment (349 words). A method \emph{strokes} on a word if its peak wrist speed within $\pm0.25$\,s of the loudness peak exceeds 1.5 times its mean wrist speed in the segment. Ground truth strokes on 171 of these words; BASE also strokes on 60\% of them and MIBURI on 43\% (the statistic quoted in \S\ref{sec:exp-decouple}). The figure shows selected cases in which ground truth and BASE stroke and MIBURI does not, at most one per speaker and chosen for being legible in still frames. Stills cannot show seams or jitter; these are shown in the supplementary video.